\documentclass[10pt,twocolumn,letterpaper]{article}
\usepackage[final]{wacv}      

\usepackage{graphicx}
\usepackage{amsmath}
\usepackage{amssymb}
\usepackage{booktabs}
\usepackage{algorithm}
\usepackage{algorithmic}

\definecolor{wacvblue}{rgb}{0.21,0.49,0.74}
\usepackage[pagebackref,breaklinks,colorlinks,allcolors=wacvblue]{hyperref}

\def\wacvPaperID{xxxxx} 
\def\confName{xxxx}
\def\confYear{2026}

\title{A Real-Time On-Device Defect Detection Framework for Laser Power-Meter Sensors via Unsupervised Learning}

\author{
\begin{tabular}[t]{c@{\extracolsep{1.5em}}c@{\extracolsep{1.5em}}c}
  Dongqi Zheng & Wenjin Fu & Guangzong Chen \\
  Purdue University & Carnegie Mellon University & University of Pittsburgh \\
  \texttt{dqzheng1996g@gmail.com} & \texttt{wenjinf@andrew.cmu.edu} & \texttt{guangzong@pitt.edu}
\end{tabular}
}
\begin{document}
\maketitle

\begin{abstract}
We present an automated vision-based system for defect detection and classification of laser power meter sensor coatings. Our approach addresses the critical challenge of identifying coating defects such as thermal damage and scratches that can compromise laser energy measurement accuracy in medical and industrial applications. The system employs an unsupervised anomaly detection framework that trains exclusively on ``good'' sensor images to learn normal coating distribution patterns, enabling detection of both known and novel defect types without requiring extensive labeled defect datasets. Our methodology consists of three key components: (1) a robust preprocessing pipeline using Laplacian edge detection and K-means clustering to segment the area of interest, (2) synthetic data augmentation via StyleGAN2, and (3) a UFlow-based neural network architecture for multi-scale feature extraction and anomaly map generation. Experimental evaluation on 366 real sensor images demonstrates $93.8\%$ accuracy on defective samples and $89.3\%$ accuracy on good samples, with image-level AUROC of 0.957 and pixel-level AUROC of 0.961. The system provides potential annual cost savings through automated quality control and processing times of 0.5 seconds per image in on-device implementation. 
\end{abstract}

\section{Introduction}

Laser power meters are critical instruments for measuring the power output of laser beams to ensure accurate and reliable laser energy measurements in medical treatments, industrial welding, and precision manufacturing applications. These devices rely on specialized sensor coatings to absorb and measure laser energy accurately. However, sensor coatings are susceptible to various types of damage during operation, including thermal damage from excessive laser exposure and mechanical scratches from improper handling.

\begin{figure}[t]
\centering
\includegraphics[width=\linewidth]{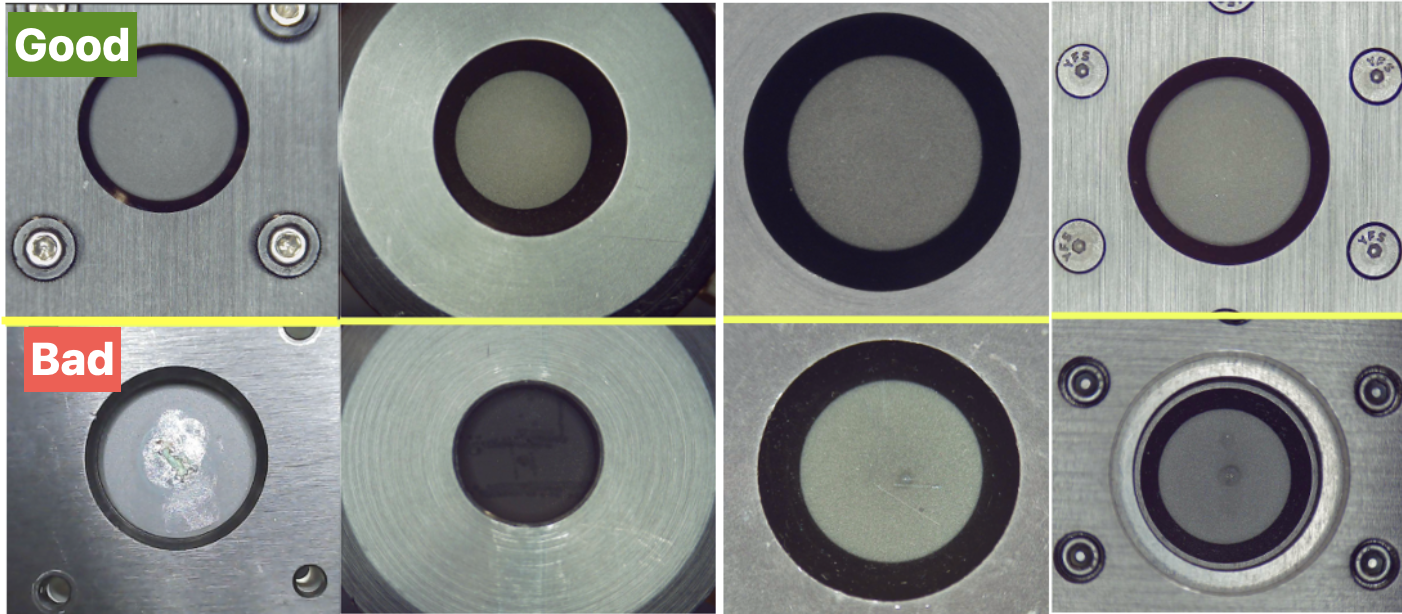}
\caption{Examples of sensor coating defects detected by quality control process. Top row shows good sensors, bottom row shows bad sensors with various defects including thermal damage and scratches.}
\label{fig:defect_examples}
\end{figure}

The current quality control process relies heavily on manual inspection on the defects (as shown in Figure \ref{fig:defect_examples}, which is time consuming, subjective, and prone to human error, which can lead to inconsistent quality decisions. Furthermore, the variety of possible defect types and their subtle manifestations make manual classification challenging even for experienced technicians.

Traditional supervised learning approaches for defect detection require extensive labeled datasets for each defect type, which are often unavailable or expensive to obtain in industrial settings. In addition, new types of defects may emerge that were not present in the training data, which limits the adaptability of the supervised methods. Our dataset consists of 366 sensor images (270 good, 96 bad) exemplifies the limited and imbalanced data typical in industrial quality control scenarios.

This paper presents an unsupervised vision system that addresses these challenges by learning the distribution of normal, defect-free sensor coatings and identifying deviations as potential defects. Our approach uses a multi-camera imaging system with sophisticated preprocessing and UFlow-based anomaly detection to achieve real-time defect classification.

\textbf{Our key contributions include:}
\begin{itemize}
\item A robust preprocessing pipeline specifically designed for circular sensor geometries that handles noise, reflections, and illumination variations through advanced circle detection and K-means segmentation
\item An unsupervised learning framework using UFlow architecture capable of detecting both known and novel defect types without requiring defect-specific training data
\item Integration of StyleGAN2-based synthetic data generation to address limited training data availability
\item Comprehensive evaluation on real industrial data achieving 93.8\% accuracy on defective samples with 0.5-second processing times suitable for production deployment
\end{itemize}

\begin{figure*}[h]
\centering
\includegraphics[width=\linewidth]{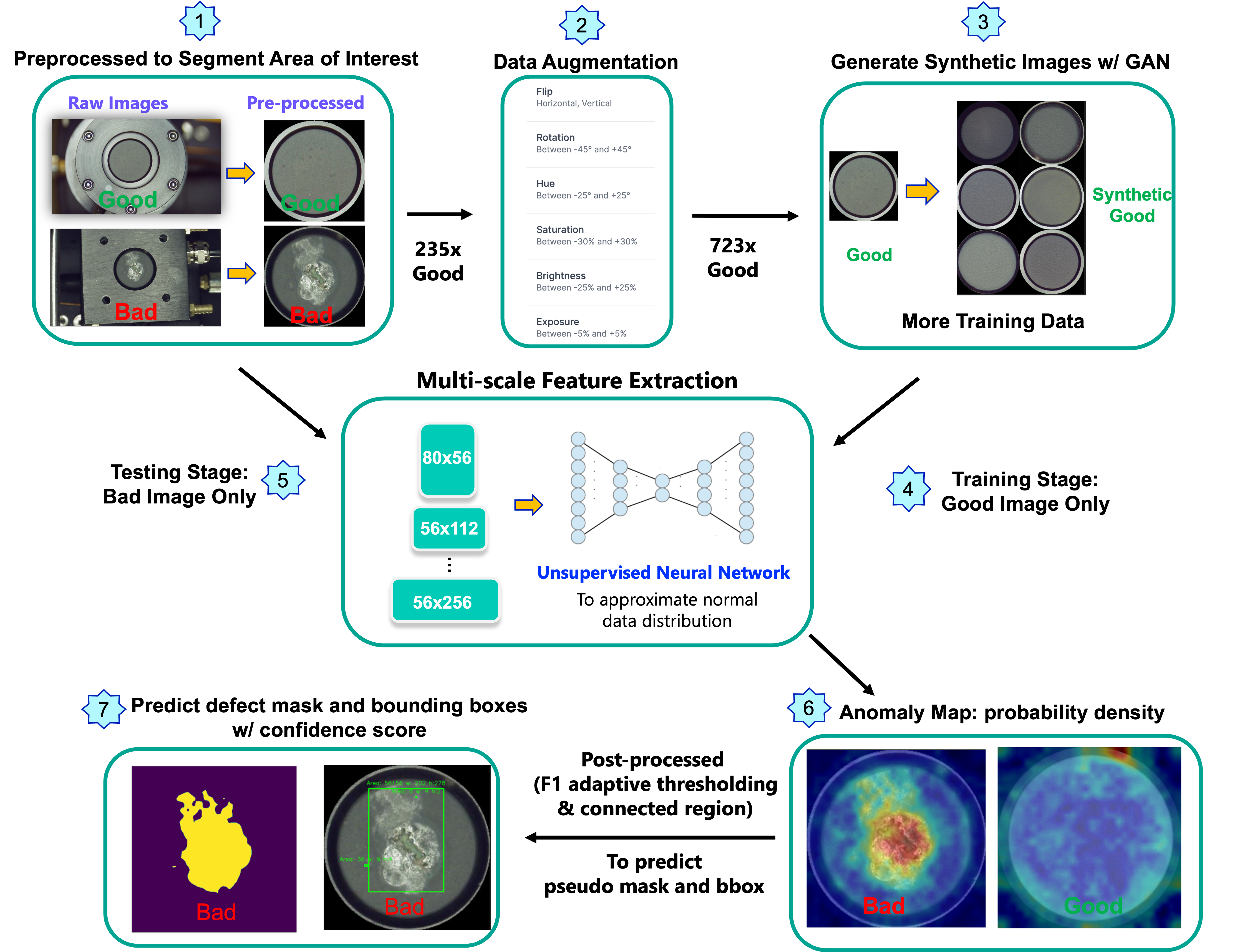}
\caption{Complete modeling framework showing the seven-stage process: (1) preprocessing to segment area of interest, (2) data augmentation, (3) synthetic image generation with StyleGAN2, (4) multi-scale feature extraction during training on good images only, (5) testing stage with bad images, (6) anomaly map generation, and (7) defect mask prediction with bounding boxes.}
\label{fig:framework}
\end{figure*}

\section{Related Work}

\subsection{Industrial Defect Detection}
Traditional approaches to industrial defect detection have relied on classical computer vision techniques such as template matching, edge detection, and statistical analysis~\cite{Newman2017}. While effective for specific applications, these methods often require extensive parameter tuning and struggle with variations in lighting, surface texture, and defect appearance.

Machine learning approaches have gained popularity in recent years, with supervised methods showing promising results when sufficient labeled data is available~\cite{Tabernik2020}. However, the challenge of obtaining comprehensive labeled datasets for all possible defect types remains a significant limitation in industrial applications.

\subsection{Unsupervised Anomaly Detection}
Unsupervised anomaly detection has emerged as a promising approach for applications where normal data is abundant but anomalous examples are rare or difficult to collect~\cite{Ruff2021}. Recent work has focused on reconstruction-based methods using autoencoders~\cite{Bergmann2018}, generative models~\cite{Schlegl2017}, and flow-based approaches~\cite{Rudolph2021}.

The UFlow model~\cite{UFlow2023} represents a significant advancement in flow-based anomaly detection, providing effective multi-scale feature extraction and probability density estimation for anomaly localization. However, its application to industrial quality control scenarios with specific geometric constraints has not been extensively studied.

\subsection{Synthetic Data Generation}
StyleGAN and its variants have shown remarkable success in generating high-quality synthetic images~\cite{Karras2019, Karras2020, Karras2021}. In the context of defect detection, synthetic data generation can help address data scarcity issues and improve model robustness~\cite{Jain2020}. However, the challenge lies in generating realistic defect patterns while maintaining the underlying normal structure. 

\section{Methodology}

\subsection{Problem Formulation}
Given a dataset of sensor images $\mathcal{D} = \{x_1, x_2, ..., x_n\}$, where $x_i \in \mathbb{R}^{H \times W \times 3}$, our goal is to learn a function $f: \mathbb{R}^{H \times W \times 3} \rightarrow \{0, 1\}$ that classifies images as either normal (0) or defective (1). The key constraint is that we only have access to normal examples during training, making this an unsupervised anomaly detection problem. Our dataset consists of 366 sensor images: 270 labeled as ``good'', 96 as ``bad''). 

\subsection{System Framework}
Figure~\ref{fig:framework} shows the full seven-stage pipeline. \textbf{(1) Preprocessing.} From an initial set of 270 verified \emph{good} images, the sensor's region of interest (ROI) is segmented (adaptive thresholding + contour filtering) and standardized via crop/resize to a fixed canvas, detailed is included in Figure ~\ref{fig:preprocessing}. \textbf{(2) Data augmentation.} The normal set is expanded to 723 images using horizontal/vertical flips, rotations of $\pm 45^{\circ}$, hue shifts of $\pm 25$, saturation changes of $\pm 30\%$, brightness changes of $\pm 25\%$, and exposure changes of $\pm 5\%$. \textbf{(3) Synthetic normals.} A StyleGAN2 generator trained on normal crops produces additional \emph{synthetic good} images to densify the normal manifold and reduce overfitting. \textbf{(4) Multi-scale feature learning (train on good only).} An unsupervised network learns the distribution of normal appearance using pyramid features at $80\times56$, $56\times112$, and $56\times256$ resolutions; no defect labels are used. \textbf{(5) Inference on suspect images.} Held-out \emph{bad} images are passed through the trained encoder. \textbf{(6) Anomaly map.} Pixel-wise anomaly scores are computed from probability density (e.g., $-\log p$) or feature-space distance to the learned normal distribution, producing a heatmap. \textbf{(7) Mask \& boxes.} F1-tuned adaptive thresholding plus connected-component analysis yields a binary defect mask; small artifacts are removed, bounding boxes are drawn around remaining regions, and a confidence score is assigned from peak/mean anomaly intensity. This design enforces strict train/test separation (training uses only normal data and synthetic normals; defects are used only for evaluation).

\subsection{Data Preprocessing Pipeline}

\begin{figure*}[!t]
\centering
\includegraphics[width=\linewidth]{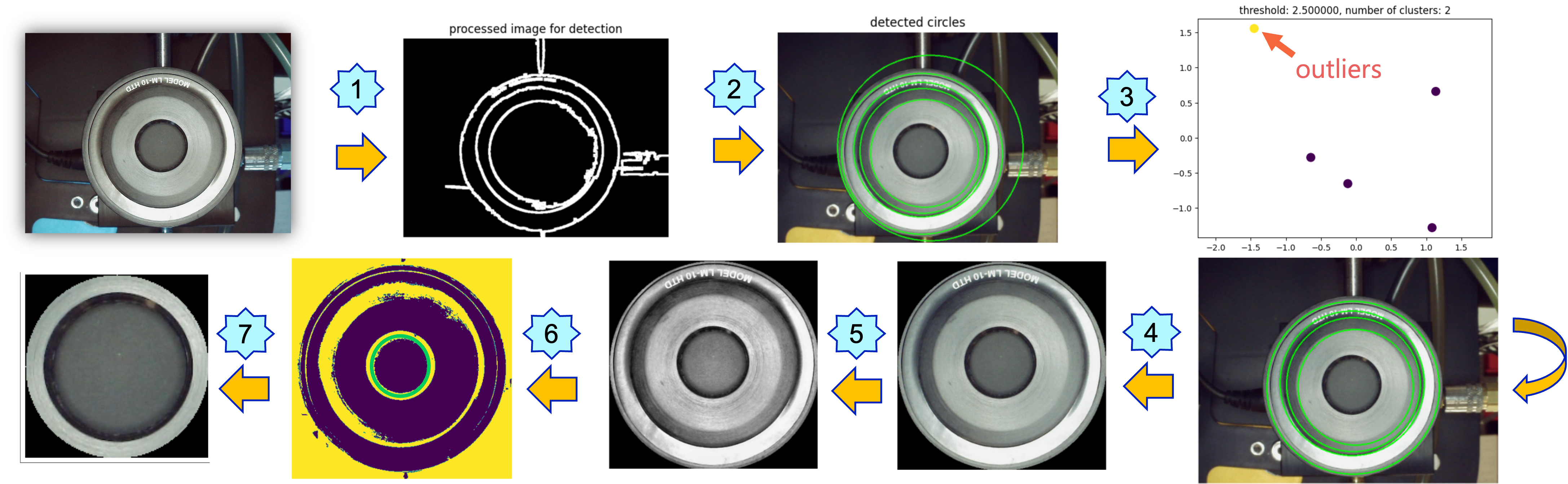}
\caption{Data preprocessing pipeline showing seven stages: (1) image processing with Laplacian edge detection, (2) circle detection using contours, (3) outlier removal, (4) cropping largest circle, (5) CLAHE enhancement, (6) K-means segmentation, and (7) final area of interest extraction.}
\label{fig:preprocessing}
\end{figure*}

The preprocessing pipeline is crucial for extracting the region of interest (ROI) containing the sensor coating while removing irrelevant background noise and artifacts. Our seven-stage preprocessing approach is illustrated in Figure~\ref{fig:preprocessing}.

\paragraph{Image Processing}
We begin by converting RGB images to grayscale and applying morphological operations to enhance edge features. Laplacian edge detection is used to identify circular boundaries, which proved superior to Canny and Sobel edge detection in our experiments:
\begin{equation}
L(x,y) = \frac{\partial^2 I}{\partial x^2} + \frac{\partial^2 I}{\partial y^2}
\end{equation}
where $I(x,y)$ is the grayscale image intensity at position $(x,y)$.

\paragraph{Circle Detection}
We employ a two-stage approach for robust circle detection. First, contours are extracted using `cv2.findContours', followed by ellipse fitting using `cv2.fitEllipse'. This approach outperformed Hough circle detection, `cv2.minEnclosingCircle', and skimage implementations due to its robustness to partial occlusions and noise.

\paragraph{Outlier Removal}
Circle candidates are filtered based on geometric constraints and distance from the image center. Outliers are removed using the Euclidean distance criterion:
\begin{equation}
d_i = \|c_i - \bar{c}\|_2
\end{equation}
where $c_i$ is the center of circle $i$ and $\bar{c}$ is the mean center of all detected circles.

\paragraph{Contrast Enhancement}
CLAHE (Contrast Limited Adaptive Histogram Equalization) is applied to the ROI to enhance local features while preventing over-amplification of noise in uniform regions.

\paragraph{Segmentation}
K-means clustering with k=2 (optimized with k-means++) is used to separate the absorber coating region from the surrounding metal ring, providing the final segmented ROI for analysis.

\subsection{Synthetic Data Generation}
\begin{figure}[!b]
\centering
\includegraphics[width=\linewidth]{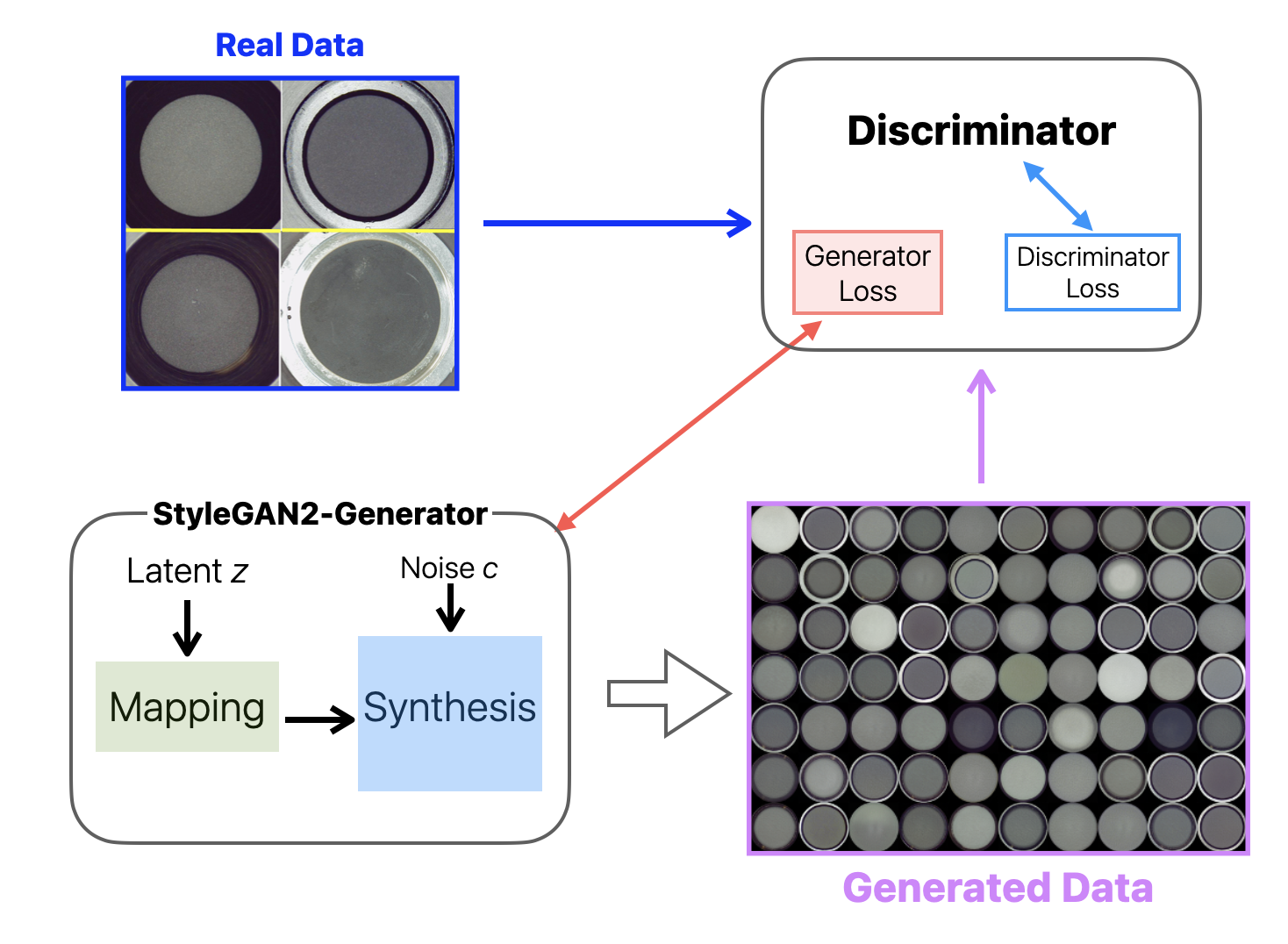}
\caption{StyleGAN2-based synthetic data generation pipeline. }
\label{fig:Style}
\end{figure}
\begin{figure}[b]
\centering
\includegraphics[width=\linewidth]{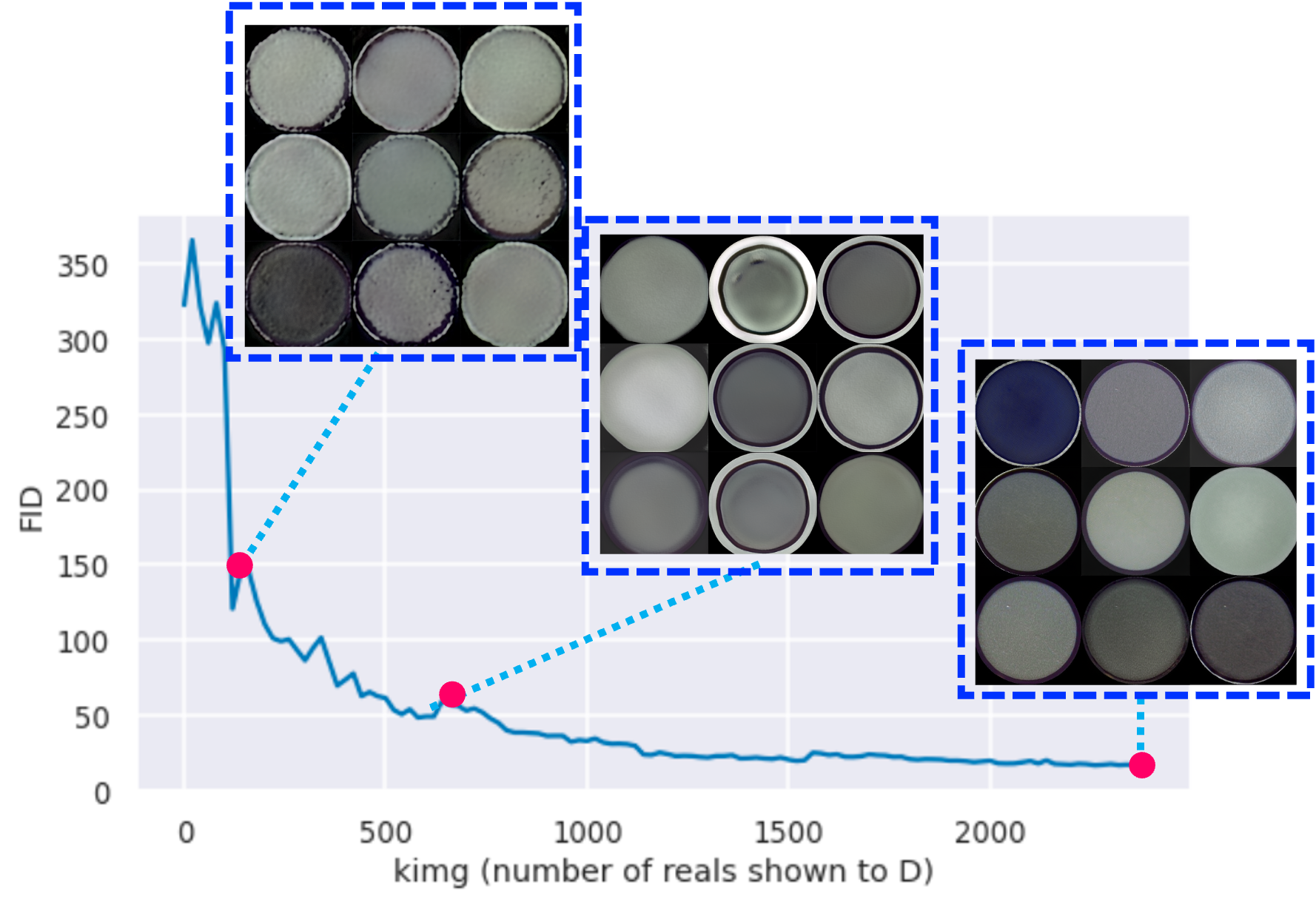}
\caption{FID plot with examples of generated synthetic images quality comparable to real samples.}
\label{fig:stylegan}
\end{figure}

To address the limited availability of training data (only 270 good images), we employ StyleGAN2 to generate synthetic images of normal sensor coatings. The approach is illustrated in Figure ~\ref{fig:Style}. Real images are used to train a discriminator that distinguishes real from fake images, while the generator learns to produce realistic sensor coating images. The plot of FID (Figure \ref{fig:stylegan} shows convergence during training, and examples of generated synthetic images demonstrate quality comparable to real samples.

\begin{figure}[b]
\centering
\includegraphics[width=\linewidth]{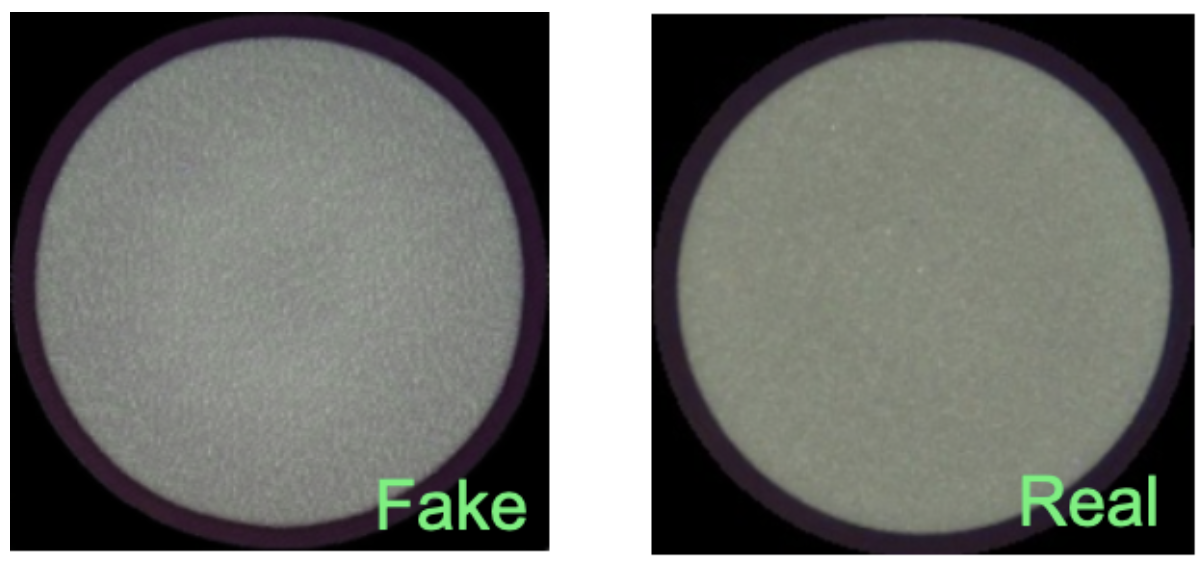}
\caption{Comparison between StyleGAN2-generated data and real data. StyleGAN2 is employed to synthesize additional ``good'' samples as a dataset augmentation method. The generated images (left) are highly realistic and almost indistinguishable from the real samples (right).}
\label{fig:fakereal}
\end{figure}

The generator network $G$ maps latent codes $z \sim \mathcal{N}(0, I)$ to realistic sensor images:
\begin{equation}
\hat{x} = G(z; \theta_G).
\end{equation}

The discriminator $D$ is trained to distinguish between real and synthetic images, while the generator learns to produce increasingly realistic samples through adversarial training:
\begin{equation*}
\min_G \max_D \mathbb{E}_{x \sim p_{data}}[\log D(x)] + \mathbb{E}_{z \sim p_z}[\log(1-D(G(z)))].
\end{equation*}

This process expanded our training dataset and provdied enhanced model robustness.

\subsection{Anomaly Detection Architecture}

Our anomaly detection system is based on the UFlow architecture, which uses normalizing flows to model the distribution of normal samples in feature space.

\paragraph{Feature Extraction}
Multi-scale features are extracted at different resolutions ($80 \times 56$, $56 \times 112$, $56 \times 256$) to capture both fine-grained texture details and global structural patterns. The feature extractor $\phi$ maps input images to feature representations:
\begin{equation}
f_i = \phi_i(x), \quad i \in \{1, 2, 3\}.
\end{equation}

\paragraph{Normalizing Flow}
For each scale $i$, a normalizing flow $F_i$ transforms the feature distribution to a standard normal distribution:
\begin{equation}
z_i = F_i(f_i; \theta_i).
\end{equation}

The likelihood of a feature vector under the normal distribution is computed as:
\begin{equation}
\log p(f_i) = \log p(z_i) + \log|\det(\frac{\partial F_i}{\partial f_i})|.
\end{equation}

\paragraph{Anomaly Scoring}
The final anomaly score combines likelihood estimates from all scales:
\begin{equation}
s(x) = -\sum_{i=1}^{3} \lambda_i \log p(f_i)
\end{equation}
where $\lambda_i$ are scale-specific weights learned during training.

\subsection{Post-Processing}
Anomaly maps are post-processed using adaptive thresholding and connected component analysis to generate bounding boxes for defect localization. The F1-optimal threshold is computed to balance precision and recall:
\begin{equation}
t^* = \arg\max_t \frac{2 \cdot P(t) \cdot R(t)}{P(t) + R(t)}
\end{equation}
where $P(t)$ and $R(t)$ are precision and recall at threshold $t$.

\section{Experimental Setup}

\subsection{Hardware Platform}
\begin{figure}[h]
\centering
\includegraphics[width=0.8\linewidth]{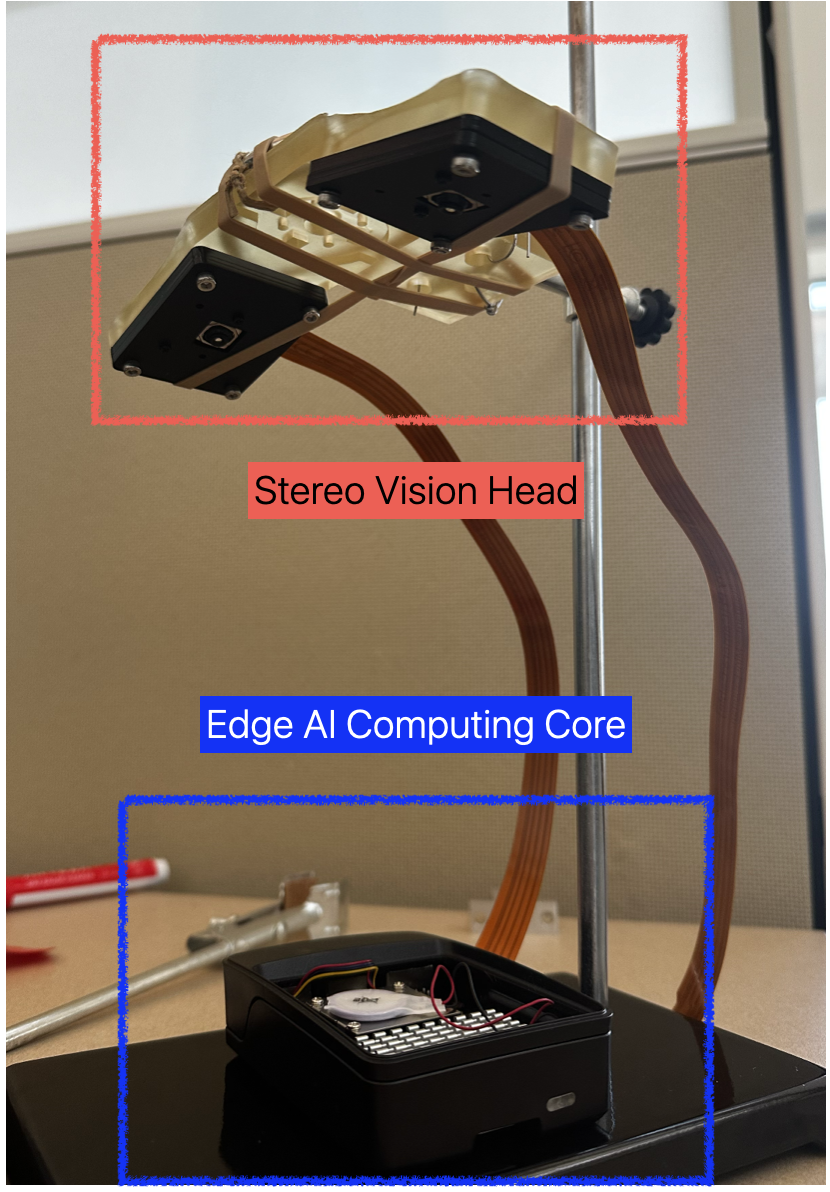}
\caption{Multi-camera defect detection system with Vision Head (red zone) containing dual cameras for stereo imaging and AI Processing Unit (blue zone) containing the Raspberry Pi-based analysis system.}
\label{fig:camera_setup}
\end{figure}

We designed a multi-camera stereo-vision system with an edge-AI computing core (e.g., NVIDIA Jetson Nano or Raspberry Pi) for on-device, real-time inference (Figure~\ref{fig:camera_setup}). The stereo head captures synchronized views from complementary angles, increasing data diversity under varied poses and illumination within a short collection window. The paired views also preserve 3D cues (depth/disparity), which we will exploit in ongoing work for geometry-aware modeling. Running inference on the edge minimizes latency and bandwidth requirements, enabling reliable, closed-loop operation in laboratory and production settings.

\begin{figure}[t]
\centering
\includegraphics[width=\linewidth]{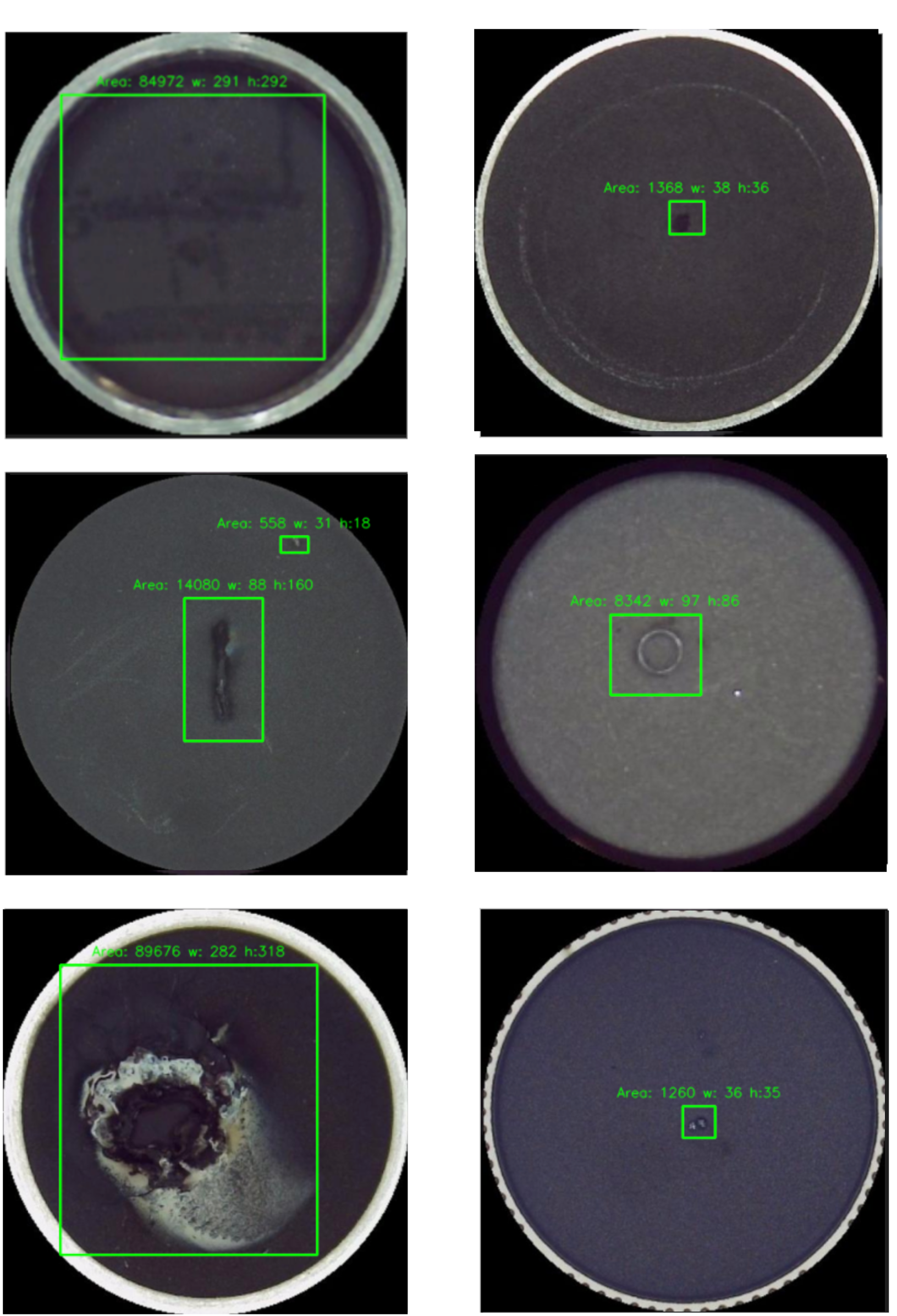}
\caption{Detection results showing successful identification of various defect types. Green bounding boxes indicate detected anomalous regions. The system successfully detects thermal damage, scratches, and other coating defects across different sensor types and lighting conditions.}
\label{fig:detection_results}
\end{figure}

Defective samples include various types of damage such as thermal damage (exposed metal), mechanical scratches, coating degradation, and other anomalies as illustrated in the detection results of Fig.~\ref{fig:detection_results}.

\subsection{Implementation Details}
The system was implemented in Python using PyTorch. The UFlow model was trained for 100 epochs with a learning rate of 1e-4 using the Adam optimizer. Orginal image went through a typical image data augmentation process including random rotations (±15°), translations (±10 pixels), and brightness/contrast variations (±25\%) to improve model robustness. The training/validation split was 80/20, with 5-fold cross-validation used for performance evaluation. Then Synthetic data generation was performed using a StyleGAN2 model trained on the normal samples for 500 epochs, expanding the training set from 723 to over 2000 good images.

\subsection{Evaluation Metrics}

We evaluate our system using standard classification metrics:
\begin{itemize}
\item Image-level AUROC: Area under the ROC curve for image classification
\item Pixel-level AUROC: Area under the ROC curve for pixel-level anomaly detection
\item Image F1-Score: Harmonic mean of precision and recall for image classification
\item Accuracy: Percentage of correctly classified images
\item Processing time: Average time per image for real-time deployment assessment
\end{itemize}

\section{Results}

\subsection{Quantitative Results}
\begin{figure}[h]
\centering
\includegraphics[width=1\linewidth]{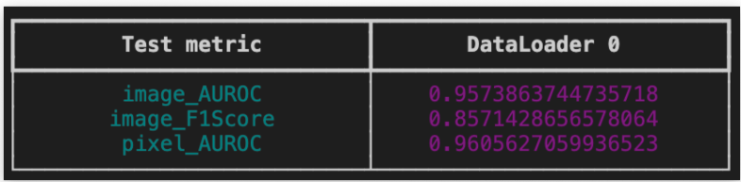}
\caption{Detailed performance metrics showing Image AUROC: 0.957, Image F1Score: 0.857, and Pixel AUROC: 0.961 achieved by our UFlow-based system.}
\label{fig:metrics}
\end{figure}

Figure ~\ref{fig:metrics} shows this system is able t achieve superior performance with an image AUROC of 0.957, pixel AUROC of 0.961, and F1-score of 0.857. 

On-device classification accuracy results show 93.8\% accuracy (90/96) on defective samples and 89.3\% accuracy (210/270) on normal samples. The average processing time is 0.5 seconds per image, making the system suitable for real-time deployment. The lower accuracy on normal samples is primarily due to subtle defects present in some images labeled as ``good'' in the original dataset, suggesting our system may be more sensitive than human inspection.

\subsection{Defect Type Analysis}
\begin{figure}[h]
\centering
\includegraphics[width=\linewidth]{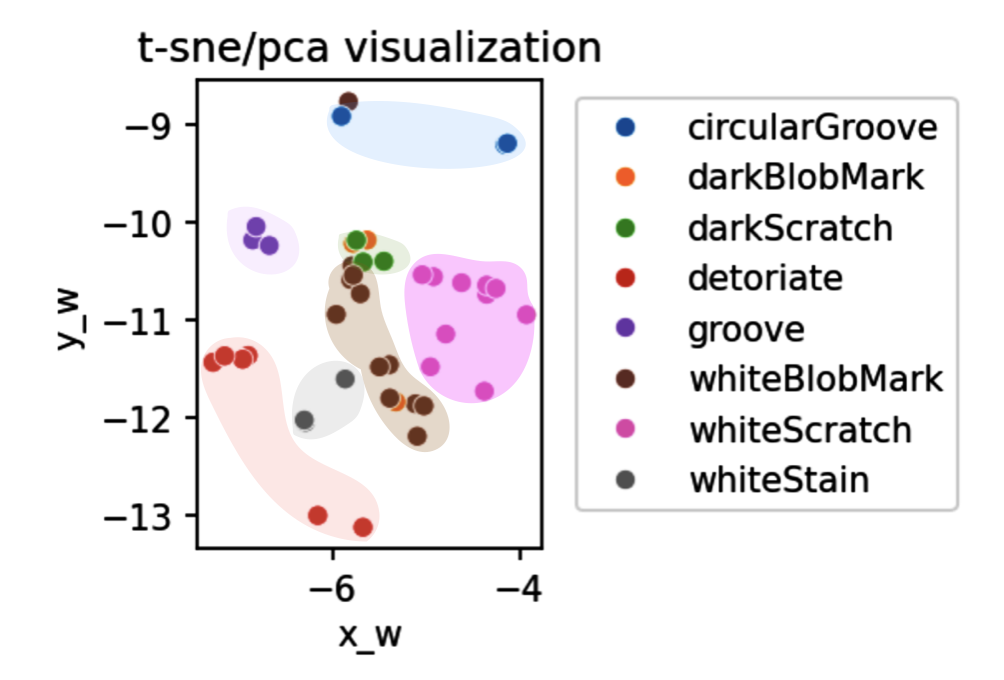}
\caption{t-SNE visualization of defect types extracted using VGG features and PCA dimension reduction. Eight distinct defect categories are identified, showing clustering patterns that could enable future defect type classification.}
\label{fig:clustering}
\end{figure}

Fig.~\ref{fig:clustering} presents the t-SNE visualization of different defect types in our dataset. We identified eight categories: \textit{circularGroove}, \textit{darkBlobMark}, \textit{darkScratch}, \textit{detoriate}, \textit{groove}, \textit{whiteBlobMark}, \textit{whiteScratch}, and \textit{whiteStain}. The clustering analysis highlights distinctive patterns among these defect types, although the limited number of defective samples (96 in total) constrains the depth of the analysis. Notably, the clusters of \textit{darkBlobMark} and \textit{darkScratch} overlap, reflecting their similar visual appearance observed in practice. This indicates that additional feature engineering or model tuning will be required to better differentiate these closely related defect types.

\subsection{Failure Case Analysis}
\begin{figure}[h]
\centering
\includegraphics[width=\linewidth]{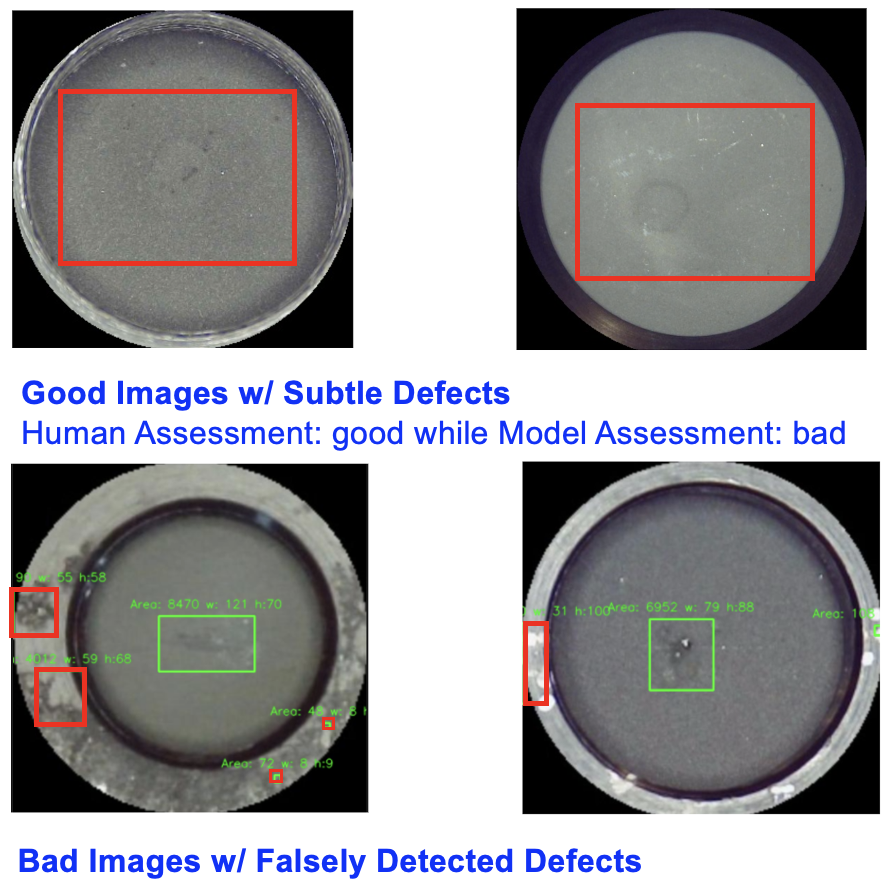}
\caption{Failure case analysis: good images with subtle defects; bad images with falsely detected defects, primarily due to edge artifacts or reflection noise. Area is marked with red frame}
\label{fig:failure_cases}
\end{figure}

Our analysis reveals two main categories of failures:\\
\textbf{(1) False Positives}
Several images annotated as ``good'' contain subtle, genuine defects that our system correctly identifies but human inspectors missed. These cases account for most of the apparent false positives, indicating that the model’s sensitivity exceeds manual inspection. As illustrated in Fig.~\ref{fig:failure_cases}, the model localizes faint defects in images labeled as good. Updating the inspection specification and relabeling guidelines to include such borderline defects would convert many of these false positives into true positives and further improve overall inspection quality.\\
\textbf{(2) False Negatives}
A small number of defective samples (6 out of 96) were not detected, primarily due to very subtle thermal damage that produces minimal visual changes in the coating appearance or defects located near the sensor edges where preprocessing artifacts may interfere with detection.

\section{Discussion}

\subsection{Industrial Impact}
The deployed system delivers tangible operational benefits for laser power-meter quality control:
\begin{itemize}
  \item \textbf{Cost reduction}: \(\sim\)\$1M annual savings is estimated at the facility from reduced manual labor and higher throughput.
  \item \textbf{Consistency}: Removes subjectivity in human inspection, yielding standardized, repeatable decisions across lots and shifts.
  \item \textbf{Scalability}: \(\approx\)0.5\,s per sensor supports in-line, real-time inspection at production speeds.
  \item \textbf{Traceability}: Every decision is logged with anomaly maps and confidence scores for auditability and corrective-action workflows.
  \item \textbf{Sensitivity}: Detects subtle defects that are frequently overlooked by manual inspection, reducing downstream field-failure risk.
  \item \textbf{Compatibility}: The multi-camera \emph{Vision Head} and \emph{AI Processing Unit} integrate with existing RMA processes with minimal workflow disruption.
\end{itemize}

\subsection{Technical Advantages}
Unsupervised anomaly detection offers several advantages over fully supervised alternatives:
\begin{itemize}
  \item \textbf{Adaptability}: Capable of surfacing novel defect types not seen during training, corroborated by the eight clusters discovered in our analysis.
  \item \textbf{Data efficiency}: Trains on normal (good) samples only, enabling deployment even when defect examples are scarce (96 bad samples in our set).
  \item \textbf{Label-noise tolerance}: Less sensitive to imperfect labels; the model often highlights subtle defects in images annotated as ``good.''
  \item \textbf{Synthetic augmentation}: \emph{StyleGAN2} increased the good-only training set, boosting robustness without additional data collection.
\end{itemize}
Last but not least, the UFlow backbone’s \emph{multi-scale} feature extraction captures both fine-grained texture and global structure—critical for coating-defect inspection—while keeping inference efficient.

\subsection{Model Selection Justification}
We benchmarked 18 anomaly-detection methods, including CFA, CFlow, CsFlow, DFKDE, DFM, and DRAEM. UFlow provided the best balance of accuracy, runtime, and stability for our circular sensor geometry and coating-texture statistics, motivating its selection as the core density model.

\subsection{Preprocessing Pipeline Impact}
The seven-stage preprocessing pipeline contributes materially to performance, yielding a 7.9\% absolute gain in image AUROC in ablations. Key elements are:
\begin{itemize}
  \item Robust circle localization via \texttt{cv2.findContours}+\texttt{cv2.fitEllipse}, outperforming Hough-based variants and \texttt{cv2.minEnclosingCircle} in our setting.
  \item Outlier suppression using center-distance heuristics to stabilize the crop.
  \item \(k\)-means segmentation (with \(k\)-means++) that cleanly separates absorber coating from the metal ring.
  \item CLAHE to normalize illumination without amplifying noise.
  \item Laplacian edges that provided more reliable circular boundaries than Canny or Sobel for these surfaces.
\end{itemize}

\subsection{Defect Type Characterization}
Unsupervised clustering (VGG features \(\rightarrow\) PCA \(\rightarrow\) t-SNE) reveals eight coherent categories: \emph{circularGroove}, \emph{darkBlobMark}, \emph{darkScratch}, \emph{deterioration}, \emph{groove}, \emph{whiteBlobMark}, \emph{whiteScratch}, and \emph{whiteStain}. While the current 96 defective samples limit fully supervised typing, these clusters provide a strong starting point for future multi-class classifiers.

\subsection{Future Applications}
Beyond the current laser power-meter deployment, the approach generalizes to advanced laser-optics inspection, AI-assisted outgoing quality control (OQC) and specification alignment, and integration with MES for fleet-wide analytics and continuous improvement, extending benefits from cost savings to systematic error reduction and data-driven process control.

\subsection{Limitations and Future Work}
Open areas for improvement include:
\begin{itemize}
  \item \textbf{Dataset scale}: The 366-image corpus constrains generalization studies and fine-grained typing; expanding cross-site data is a priority.
  \item \textbf{Label inconsistencies}: Some ``good'' images harbor subtle defects; specification updates and human-in-the-loop relabeling will reduce apparent false positives.
  \item \textbf{Subtle defects}: Six false negatives indicate room to improve sensitivity to low-contrast thermal damage (e.g., via higher-resolution crops or adaptive thresholds).
  \item \textbf{Edge cases}: Defects near sensor boundaries can be impacted by cropping; boundary-aware preprocessing will mitigate this.
  \item \textbf{Real-time optimization}: Although 0.5\,s/image meets current needs, further pruning/quantization would raise line throughput.
  \item \textbf{Noise robustness}: Glare, reflections, and color shifts remain challenging; domain-randomized augmentation and photometric normalization are planned.
  \item \textbf{Human feedback}: Operator-guided threshold tuning (score-histogram calibration) to refine \emph{good/OK/bad} decision boundaries.
  \item \textbf{Defect typing}: With 96 defective samples across eight clusters, scaling labeled data is essential for reliable supervised type classification.
\end{itemize}

\textbf{Future work} will expand data across manufacturing sites; incorporate operator-in-the-loop threshold calibration via score-histogram feedback to refine decision boundaries; develop robust, supervised defect-type classifiers as labels scale; and extend the approach to additional optical-component QC tasks. The successful mass-production deployment establishes a foundation for broader, global roll-out of industrial vision systems.

\section{Conclusion}
We present an end-to-end, vision-based system for automated defect screening of laser power-meter sensors using \emph{unsupervised anomaly detection}. The method couples a seven-stage preprocessing pipeline tailored to circular sensor geometries with a UFlow-based density model and multi-scale feature extraction. To mitigate limited data, \emph{StyleGAN2} augmentation expanded the good-only training set with traditional data augmentation methods. The system achieves \textbf{0.957} image AUROC, \textbf{0.961} pixel AUROC, \textbf{0.857} F1, and \textbf{93.8\%} accuracy on defective samples, with an inference time of \textbf{0.5~s/image} suitable for production. In deployment, it promises substantial operational impact by avoiding manual-inspection costs on site---while improving consistency over human inspection.

Architecturally, a multi-camera setup with a \emph{Vision Head} and \emph{AI Processing Unit} integrates cleanly with existing RMA workflows. Unsupervised defect clustering (VGG features + t-SNE) reveals eight coherent families---\textit{circularGroove}, \textit{darkBlobMark}, \textit{darkScratch}, \textit{deterioration}, \textit{groove}, \textit{whiteBlobMark}, \textit{whiteScratch}, and \textit{whiteStain}---laying groundwork for future, fine-grained defect classification, indicating higher sensitivity than manual inspection and the potential to raise quality standards.

{\small
\bibliographystyle{ieee_fullname}

}

\end{document}